\documentclass[compsoc,conference,a4paper,10pt,times]{IEEEtran}
\usepackage{graphicx}
\usepackage{booktabs}
\usepackage{multirow}

\usepackage{todonotes}
\usepackage{url}
\usepackage{hyperref}
\AtBeginDocument{%
  }

\begin{document}

\title{Towards Improved Anomaly Detection for Cloud Cybersecurity via Graph Neural Networks}

\author{
\IEEEauthorblockN{Manu Nandan}
\IEEEauthorblockA{\textit{CrowdStrike, Inc.}, USA\\
manu.nandan@crowdstrike.com}
\hfill
\vspace{1em}
\IEEEauthorblockN{Michael Brautbar}
\IEEEauthorblockA{\textit{CrowdStrike, Inc.}, USA\\
michael.brautbar@crowdstrike.com}
\and
\IEEEauthorblockN{TJ Jaymes}
\IEEEauthorblockA{\textit{CrowdStrike, Inc.}, USA\\
tj.jaymes@crowdstrike.com}
\hfill
\vspace{1em}
\IEEEauthorblockN{Edward Raff}
\IEEEauthorblockA{\textit{CrowdStrike, Inc.}, USA\\
edward.raff@crowdstrike.com}
\IEEEauthorblockA{\textit{Univ. of Maryland, Baltimore County}, USA\\
raff.edward@umbc.edu}
}

\maketitle

\begin{abstract}
  Detecting security threats in an organization's cloud computing environment has become necessary due to the increased reliance on cloud infrastructure. Logging of all cloud computing events enables investigation into any incidents after they are detected. Automated detection of threats using the logs based on heuristics or anomaly detection could result in a high false positive rate due to its relatively static nature. In this article, we present an industrial case study of a self-supervised learning method using graph neural networks applied to AWS CloudTrail logs to surface suspicious events for analyst review. The model produces an anomaly score for each event and dynamically adapts to changes in the organization without requiring periodic retraining. Based on our experiments across five organizations, the proposed model produced substantially fewer alerts than a domain expert rule-based baseline in almost all cases, reducing alert volumes to approximately 1 per hour from thousands generated by traditional methods. \textbf{We note that this evaluation covers only flagged events, and false negatives cannot be estimated from the current data; findings should therefore be interpreted as a practical deployment study offering insights into real-world constraints rather than a fully validated detection system. We discuss these limitations and the requirements for extending the approach to other cloud environments as future work.}
\end{abstract}


\section{Introduction}
Modern day organizations typically use significant computer assets on the cloud, such as AWS, GCP, Microsoft Azure, etc. Any access of cloud resources is logged to enable housekeeping, identify errors, and investigate any malicious activity. For example, AWS CloudTrail logs provide details of each API call made from accounts owned by an organization, enabling identification of inefficient usage of AWS resources, identification of an ongoing cyberattack, or investigation of the origins of a cyberattack post-event. While such logs hold valuable information for detection and remediation of lapses in cloud security, their huge volume can make it very mundane and time-consuming for security analysts to routinely study them for the purposes mentioned above. It is common for organizations with a large cloud footprint to have millions of events logged each day. Typically, analysts rely on heuristics to identify suspicious events that they can investigate further. For example, a user assuming an admin role for the first time to access privileged data can warrant an investigation. While such rule-based methods can work efficiently in some well-known cases, they can miss a wide variety of events that might lead to breaches due to the dynamic nature of events. In addition, rule-based methods require experts to maintain them, which can be expensive and error-prone. While combining such approaches with anomaly detection based methods can improve performance they can also be slow to adapt or expensive. Any automation to help reduce the manual efforts of analysts and improve breach detection is of great value. 

This article describes our research into identification of anomalous events in AWS CloudTrail logs with a focus on industrial applications. Due to the simple nature of the features extracted from the events, the general strategy is expected to have applicability to other cloud logs. Considering the large variety of events and the fact that each organization's security posture and concerns are different, a supervised learning approach is not feasible. We also describe our research on applying a self-supervised learning approach to this problem, and as shown in the results section, we observed a ten-fold increase in the number of events surfaced that were of medium risk or higher, compared to detections by a baseline approach among flagged events. Specifically, we used Temporal Graph Networks (TGNs) \cite{Rossi20} a Graph Neural Network (GNN) that is very well suited for this application. GNNs, by design, excel at processing data that is in a graphical format. They can capture the essence of relations between entities in graphs without the need for sophisticated hand crafted features. The result is that they have exceptional performance and are widely used in the industry in several applications \cite{Sharma24,Ying18, Austin21, Sha25, Wu24}.  Real world challenges of evaluating and adapting such a method for use in industry along with the limitations of our approach are described in Section \ref{sec:limits}.

\begin{figure}[h!]
\centering
\includegraphics[width=0.75\linewidth]{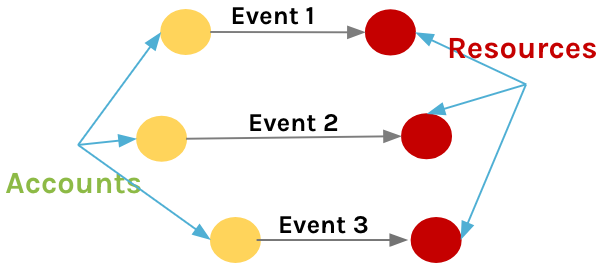}
\caption{Representation of CloudTrail logs as a graph. The nodes on the left represent accounts, while nodes on the right represent resources. Events initiated by accounts on resources are represented as edges linking their nodes.}
\label{fig:base_graph}
\end{figure}

CloudTrail logs are naturally well suited to be represented as a graph. Each event is essentially a link from an account to a resource or in other terms an edge from an account node to a resource node. An example of such a graph is shown in Fig. \ref{fig:base_graph}, where the nodes are the accounts or cloud resources. The edges in this graph represent an event connecting them, and details of the event can be represented as edge features. The method considers minimal processing of data to preserve some of the domain expertise in the field, instead of being completely domain agnostic as in \cite{Farzad20, Min17}. While GNNs have been previously used in cybersecurity applications, this is the first study on using them specifically for threat detection on cloud logs to the best of the authors' knowledge. In addition, customizations of the specific GNN training process, which are designed to work well with such data, are also proposed.

Compared to anomaly detection methods, our solution has the following operational advantages:
\begin{itemize}
\item Adaptability during inference: the nature of an anomaly changes over time. Organizations have accounts or resources added or removed dynamically. Our solution adapts to changes in the environment easily without requiring frequent model retraining.
\item Anomalies are by design identified at an account or resource level due to the nature of the underlying algorithm, instead of an organization wide definition, which in our experiments resulted in fewer false positive alerts compared to a rule-based anomaly detection baseline. Further experimentation is needed to analyze impact on missed detections (False Negatives) as described in Section \ref{sec:limits}.
\end{itemize}

\section{Related Work}

The database community has increasingly been studying issues in cybersecurity due to the interaction of graph structures that are extractable in columnar and graph-based databases. Most recently \cite{song_advancing_2025} exploited graph structures and neural networks to classify an attack into one of four different types to aid security analysts. Similarly, ~\cite{king_trail_2025} uses graphs to identify new Indicators of Compromise (IoC) for a \textit{known} attack. However, neither work details how to identify the attack first -- which we study in this article. Such graph representations are also used to support reverse engineering~\cite{saqib_gage_2024} and report generation~\cite{gao_query_2019,gao_system_2021} help inform the useful features for detection, but are beyond the scope of this work. 

Anomaly detection on CloudTrail logs using machine learning has been researched \cite{Hagemann21} previously, with most of the solutions requiring feature engineering to capture the topological information in the graph of Fig. \ref{fig:base_graph}. While Large Language Models (LLMs) are being studied, their computational cost makes them infeasible for financial and latency reasons at this time~\cite{sui_bridging_2025}. Non-LLM solutions that apply graph-based methods typically consider static graphs \cite{Abdullayeva24, Nguyen22, Xu24, Elsayed20} on data such as network flow logs. Static graphs were the focus of most early research on GNNs, where only edge features change with time, like in the example of traffic prediction for Google Maps \cite{Austin21}. Dynamic graphs can change their topology with time by either changing node features or even adding and removing nodes, for e.g., graphs of users in social media. The recent research on dynamic graphs is of interest to cybersecurity in general because most of the use cases in the domain seem to require dynamic graphs. Unlike the commonly used GNN methods such as GCN \cite{Kipf16}, TGN \cite{Rossi20} works on dynamic graphs where nodes can be added or removed as needed. For example, the addition of new accounts or resources such as an S3 bucket would be gracefully handled by the model with the addition of new nodes. Additionally, it is crucial to consider the time-based cloud resource usage patterns. Not only is it helpful to find events from user accounts that occur at atypical times for that account, but it is also useful to identify cases of evolution of usage patterns of a group of similar accounts. The temporal evolution of dynamic graphs is also represented in TGN as events are processed in chronological order. There are two approaches to modeling dynamic graphs: 

\begin{itemize}
\item Discrete Time Dynamic Graphs (DTDG): Works on sequences of static graph snapshots taken at fixed intervals of time. One such solution is to divide a temporal graph of network flow data into snapshots taken over time intervals to train a combination of GNN and RNN \cite{King23} for intrusion detection.
\item Continuous Time Dynamic Graphs (CTDG): Does not need data to be in the form of snapshots at fixed intervals, but rather processed as a sequence on events \cite{Xu2020}
\end{itemize}

TGNs build upon several other GNN models that can work on CTDGs and is the most general of them. A particularly useful feature is its ability to handle edge features. It has been reported to give the best performance \cite{Shenyang23, Hongkuan22} on the largest of the graph datasets. TGNs follow the message passing neural network framework \cite{Gilmer17}. TGNs are trained on unlabeled data using self-supervised learning, where existing edges are assumed to have a target value of 1 and generated negative samples are assumed to have a target of 0. During inference the model outputs a score via a sigmoid that reflects the likelihood of a given edge of specified features existing between the nodes; in this work we treat this output as an anomaly score.

\subsection{Temporal Graph Network Model} \label{sec:tgn_def}
\begin{figure}[h!]
    \centering
    \includegraphics[width=1\linewidth]{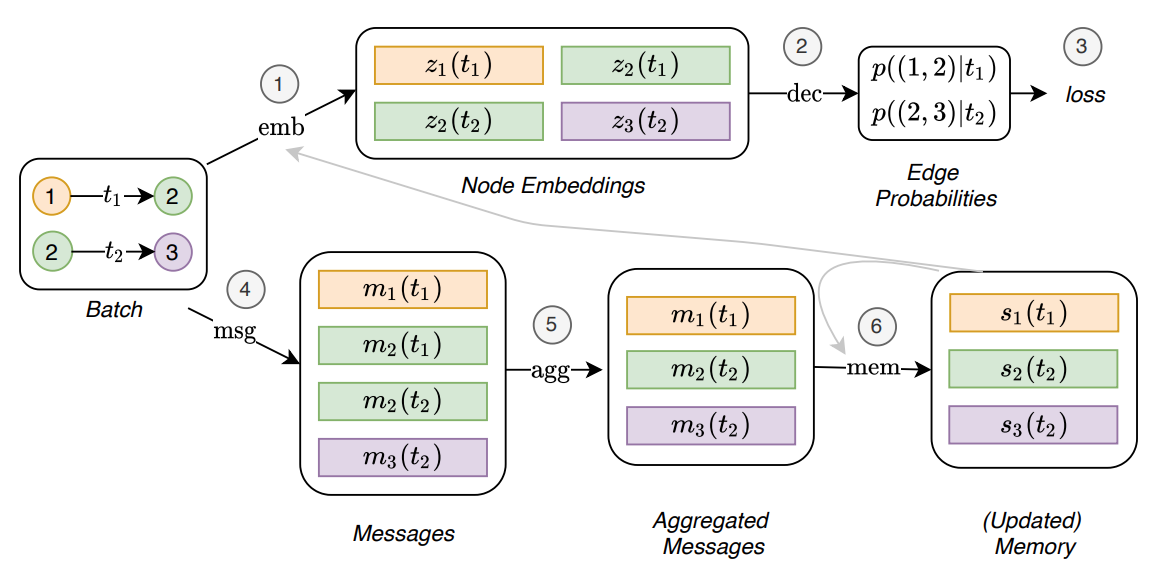}
    \caption{Computational blocks in a TGN model from \cite{Rossi20}. The sequence of events are processed in batches, where each event is an edge in the graph. The model computes an anomaly score for each edge based on embeddings of the related nodes. The embeddings of each node are updated based on the stored 'memory' of the same nodes and their neighbors.}
    \label{fig:TGN}
\end{figure}

Fig. \ref{fig:TGN} illustrates the building blocks of the TGN model and how it processes a batch of event data for model training. Given $s_i$ and $s_j$, the memory of nodes $i$ and $j$, that have an event $e$ at time $t$, the message $m_i$ from node $i$ = $msg(s_i , s_j , t, e)$. The message function can be computed using any function such a combination of input features or multi-layer perceptrons (MLPs). A message aggregator combines multiple events linked to each node using an aggregation function such as the mean of the messages of that node. Finally, the update function uses the aggregated node message and current node memory to compute the new node memory using gated recurrent units (GRU). This is another practical advantage of TGN, where the model updates its memory during inference, allowing future predictions to be based on events that occur after the training period. The node embeddings for node $i$, at time $t$ are computed using graph attention applied to previously computed embeddings of node $i$ and its neighbor nodes and of the edges connecting them. The embeddings are fed to a decoder (MLP) that generates an anomaly score for each edge via a sigmoid output in the range $(0, 1)$. Note that because TGN is not a Bayesian model, this output is not a calibrated probability; in this work we treat it as an anomaly score and apply a threshold to flag events for review, consistent with standard practice in the anomaly detection literature. For training the model, the binary cross-entropy loss function is used.

\section{Method}
\subsection{Data Processing}
Cloud logs capture all events within a customer's environment. This could include any API calls from services accessing their cloud resources, API calls from SDK or CLI commands by human users. Organizations with good security posture are found to have a very different set of cloud logs compared to others with extensive use of automation, limiting access to a few roles, etc. An example is the automation of GitLab runners that is necessary for large organizations to manage CI/CD. The usage of resources can also vary widely, for example an organization that sells fitness trackers will have a constant stream of updates of data in S3 in contrast to retail stores that might have periodic updates of data. We use simple feature engineering to create a normalized set of features that can be used with all customers. In addition, the features are designed to indicate possible security risks rather than capturing all information about the event logs.

\subsubsection{Feature Extraction} \label{sec:feature_extraction}

\begin{figure}[h!]
    \centering
    \includegraphics[width=0.7\linewidth]{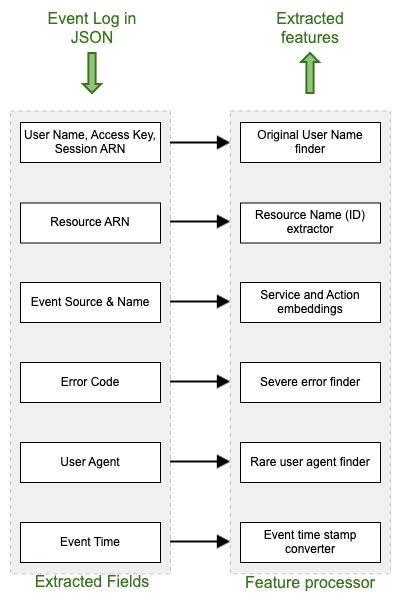}
    \caption{Method of feature extraction from raw event logs that is used in this study. The features are generic so that it is easy to adapt to other cloud providers. They represent high-level information of the events with a focus on some of the indicators of malicious activity that a cybersecurity analyst typically looks for.}
    \label{fig:features}
\end{figure}

From each log, the following information is extracted as shown in Fig. \ref{fig:features}:
\begin{itemize}
    \item Account linked to the event: As described in the event account identification section, this will be the original AWS account that directly or through an assumed role initiated the event. When the original account cannot be traced from the available data, the principal ID associated to the event log is used.
    \item Resource linked to the event: The resource names are specified in many events as part of a resource ARN (Amazon Resource Names). However, there are events where the resource name is not specified in this format. We focused on the fully qualified ARN in the top level of the event log. Any event where such a resource ARN (at the top level of the log) was not provided was ignored. Though this resulted in the loss of some data, we believe we have a sufficient number of logs to demonstrate the feasibility of the approach. From the resource ARN, the resource id is extracted based on the ARN format specified by AWS. The resource id is further processed for logs of s3 service, where only the bucket name is retained. For example, if the s3 location mentions the folder structure and file name, all that information is ignored and only the bucket name is retained. This process ensures that the relevant information is retained and we can identify multiple access to the same bucket even if they are to different sub-folders within it.
    \item AWS service of the event, with more details in \ref{sec:embd_aws_service}.
    \item Action within the service of the event, with more details in \ref{sec:embd_aws_action}.
    \item Errors in the event: The error code logged in some CloudTrail logs carries vital information, such as when it is a 403 exception ('Access Denied') rather than a 400 exception ('Bad Request'). Using AWS common error types we can get the mapping from exception type to HTTP type such as 'ForbiddenException' to 403 or 'NoSuchWebsiteConfiguration' to 404. Any error that is of type 403 or 401 is considered as a boolean feature 'Has Severe Error'. The presence of any error is also a boolean feature: 'Has Error'.
    \item User agent type: The user agent type is also of interest from a cybersecurity standpoint, especially when it is of a rare type. Any user agent type that, on average, is used less than once per hour is considered rare in this research, yielding the feature 'Is Rare User Agent'. In addition, user agents that are used in less than 10$\%$ of the events (in the training set) are identified with the feature 'Is User Agent Less Than 10p'.
    \item Event time: The time of the event is intrinsically considered in the TGN model.
\end{itemize}

\subsubsection{Event Account Identification}
The CloudTrail logs need to be parsed using logic that incorporates knowledge of their format. Identification of the actual account responsible for an event is not simple. It is a common practice for AWS users to first assume a role and then initiate events with the assumed role. When a user assumes a role through an AssumeRole or similar event, that event’s log will have an access key and a session token that can be used to identify the user. All such events need the processing of access keys and session names to identify the associated accounts. For this study the parser does such processing and attempts to identify the original account using the access keys and session names.

\subsubsection{Embeddings Of AWS Service} \label{sec:embd_aws_service}
Risk profiles of different AWS API's are well studied in the open source community, with information available about them widely \footnote{For our study, the following package was used \url{https://github.com/iann0036/iam-dataset/tree/main/aws/managedpolicies}}. Using this information, the number of actions that can lead to the following types of security risks is calculated for each service:
\begin{itemize}
\item Data access provision
\item Privilege escalation
\item Credentials exposure
\item Resource exposure
\end{itemize}
The resulting counts for each of the four categories of each service is scaled to the range (0,1) and used as the embeddings of that service. In addition, the services are grouped based on how common or rare related events are found to be in the logs of each customer. For each customer we get the counts of events of each service and order them in decreasing order. The services are then grouped based on percentile ranking as group 1: up to 60$\%$, group 2: 60 - 80$\%$, group 3: 80 - 95$\%$, and group 4: 95$\%$ and more. Thus, altogether the service of each event is represented by eight features.

\subsubsection{Embeddings Of AWS Service Action} \label{sec:embd_aws_action}
Using the same information as described above, we represent each service action (API) by whether they represent security risk in the four categories. In addition, the prefix of the action is used to identify if it represents a security risk. For example, in AWS the prefix typically conveys meaning about the nature of the action. The ten action prefixes with the most possible security risks (similar to the calculation described above for the AWS services) were identified. Any event with an action that has one of these prefixes is given a value of one for the feature ('is risky action prefix'). This is helpful to make the features work with even actions that are newly added to AWS. Thus, in total five features represent the embeddings of the action.

\subsection{Modifications To Self-Supervised Learning For Link Prediction} \label{sec:mod_self_super}
In this study, we use the model to identify anomalous and possibly malicious events using its link prediction ability. Specifically, the TGN model is trained with a self-supervised learning approach to predict a label of 1 for existing edges and 0 for non-existing (unlikely or anomalous) edges. Since the model sigmoid output is a continuous value in the range (0,1), it will compute a score close to 1 for edges similar to those seen previously and a score close to 0 for those that are not similar to previously seen edges. We treat this output as an \emph{anomaly score}, and apply a threshold to flag events for analyst's review. This process eliminates the need for labeled data and hence makes a per customer model practical.

\begin{figure}[h!]
    \centering
    \includegraphics[width=0.8\linewidth]{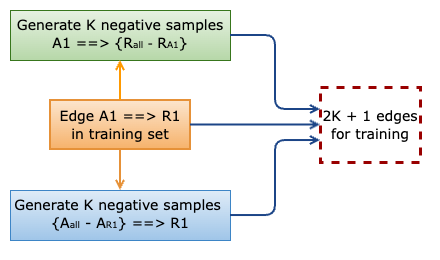}
    \caption{For each edge from account node $A1$ to resource node $R1$ in the training set, multiple negative samples are generated. $R_{all} - R_{A1}$ indicates resources that do not have any edge to $A1$ and $A_{all} - A_{R1}$ are account that do not have any edge to $R_1$ in the training set. This scheme results in a more representative negative samples compared to just randomly generated negative samples.}
    \label{fig:negative_sampling}
\end{figure}

Some modifications to the self-supervised learning process were necessary for the use of TGN in this project. As mentioned earlier, TGN training is self-supervised, where existing edges are used as positive training samples and randomly generated non-existent edges are used as negative samples. A slight tweak of this process is needed to make it work well for event logs. Consider two nodes, an account node ${A_1}$ and a resource node ${R_1}$. Negative samples should represent non-existent edges from ${A_1}$ to other resources $R_{all} - R_{A1}$, where $R_{all}$ represents the set of nodes of all resources and $R_{A1}$ denotes the set of all resource nodes that have edges to $A_1$. The negative samples should also include non-existent edges from other account nodes $A_{all} - A_{R1}$ to resource ${R_1}$, where $A_{all}$ represents the set of nodes of all resources and $A_{A1}$ denotes the set of all resource nodes that have edges to $A_1$. Hence, every positive edge has 2K corresponding negative edges as shown in Fig. \ref{fig:negative_sampling}. The loss function used during training, the binary cross-entropy loss, is weighted to give a weight of K for the loss due to negative samples. Note that $R_{all}$, $R_{A1}$, $A_{all}$, and $A_{R1}$ are computed using only the training data during model training.

Another modification to the self-supervised learning process was needed to improve the detection of possibly malicious activity. Consider the example where for one user account, several events in the training data have errors, such as 'Access Denied', which can indicate malicious activity. A burst of such activity might happen during training by chance, but any method learning from only on a chunk of data with such activity will consider it to be routine. The TGN model will also consider such events as typical of positive edges and hence compute a high anomaly score for them during inference, leading to false negatives. The modification was to consider well-known indicators of malicious activity in events in the training data to make the model predict low score for possibly malicious events, irrespective of how common such events are in the training data. During training, the target label for such events is reduced depending on the severity of the error so that the model learns to identify such events as rare. An extension was to lower the target score for events linked to unlikely user agents, as that is also an indicator of possibly malicious activity.

In addition to the modification to negative sampling, there also needs to be a corresponding change in the loss function used during training. Typically, it is sufficient to add the mean loss from prediction for actual edges and negative sample edges for each batch as the training loss of that batch. However, weighting the mean loss from negative sample edges by K was found to yield better results.

\subsection{Graph Representation Of Data}
\begin{figure}[h!]
    \centering
    \includegraphics[width=0.75\linewidth]{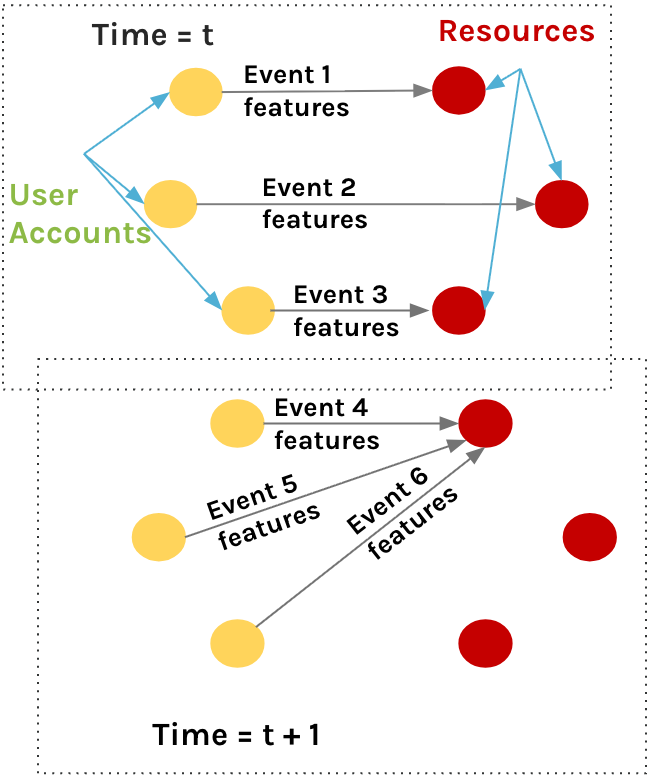}
    \caption{Graph representation of model A with a focus on relationship between accounts and resources. All event information described in Section \ref{sec:feature_extraction} is used to construct this graph. The dynamic nature of the graph is represented by the two dotted rectangles showing events at time $t$ and one second after $t$.}
    \label{fig:graphA}
\end{figure}

The accounts and resources can be considered as nodes in a graph with the features mentioned above forming their edge features. The time feature is considered with a resolution of one second. The graph shown in Fig. \ref{fig:graphA} is used to train the primary model (model A) of this research. The purpose of this model is to predict the likelihood of any given event, considering its account, resource and all other features described in Section \ref{sec:feature_extraction}.

While this might seem sufficient, it was found that organizations can routinely create new resources that can lead to a burst of spurious alerts. Filtering out bursts of alerts might be possible to an extent with rule-based approaches. However, for this research, we used a second graph representation of the data (Model B) that can learn the relation between users and events without considering the resources. Model B is used to determine if it is unlikely that the user experienced any events of the observed type. The edge features are a subset of those used for model A: based on user agent, error codes, and time.  Using model B is expected to provide a more adaptive method of filtering out spurious alerts compared to a rules-based approach, since its predictions will be influenced by patterns of other users also.

\begin{figure}[h!]
    \centering
    \includegraphics[width=0.75\linewidth]{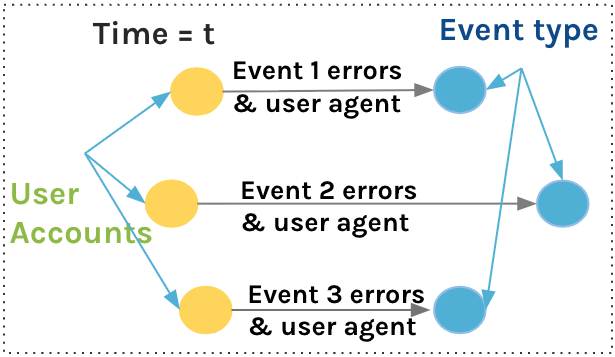}
    \caption{Graph representation of model B with a focus on relationship between accounts and event types. The resource associated with the event is ignored. The only edge features are those about event errors and user agent described in Section \ref{sec:feature_extraction}.}
    \label{fig:graphB}
\end{figure}

\section{Results}

\subsection{Datasets}
CloudTrail logs from 5 different organizations of varied size were collected over a period of 60 days for experimentation, as shown in Table \ref{tab:dataset_stats}. Model training was done on 95$\%$ of the data and testing on 5$\%$, which corresponds to the logs of roughly the last 3 days. The predictions for the test set were analyzed in detail by a subject matter expert in CDR (cloud detection and response). For the purpose of comparison, a baseline rule-based approach was also implemented. The baseline detector only identifies events that are unusual for an account–resource pair or events that have indicators of malicious activity, such as the presence of errors, as described below. The alerts shortlisted by the model are those that are found to have an anomaly score $\leq$ 0.5 for model A and 0.75 for model B. Note that training was done using 1 GPU of a g5.12xlarge EC2 instance for each dataset, while testing was done using CPU only. For a batch size of 16 during inference, the time taken for prediction on CPU is less than 1 millisecond per event, measured only for the de-duplicated events. Events are deduplicated as, those that have identical account, resource, event source and name (service $\&$ action), user agent, and any errors. In addition, the event time is considered in 30 minute buckets. 

\begin{table}[htbp]
\centering
\caption{Datasets used for evaluation. All data was collected over 60 days.}
\label{tab:dataset_stats}
\setlength{\tabcolsep}{3pt}
\begin{tabular}{c|cc|cc}
\toprule
\multirow{2}{*}{\textbf{Dataset}} & \multicolumn{2}{c|}{\textbf{Data Size}} & \multicolumn{2}{c}{\textbf{TGN Training}} \\
\cmidrule(lr){2-3} \cmidrule(lr){4-5}
& \textbf{Events} & \textbf{Deduped} & \textbf{Epochs} & \textbf{  Time per Epoch (s)} \\
\midrule
1 & 4.9M & 437K & 6 & 185 \\
2 & 20.7M & 522K & 6 & 214 \\
3 & 13.3M & 286K & 6 & 114 \\
4 & 38.5M & 18.1M & 3 & 5K \\
5 & 12.5M & 5.8M & 3 & 1.6K \\
\bottomrule
\end{tabular}
\end{table}

\subsection{Baseline Rule-Based Approach}

For the baseline detector, the following counts were computed for each user account from the training data:

\begin{itemize}
    \item Counts of events per resource.
    \item Counts of events types including service and action.
    \item Counts of service types.
    \item Counts of user agent types: number of times a user agent was typical (used more than once per hour on average) and number of times it was not.
\end{itemize}

For each event in the test data, the statistics above are used to compute an anomaly score as described below. Events with score $\geq 0.5$ are reported as anomalies.

\begin{itemize}
    \item For the event, add a weight to the score for each category if the count of that type is less than the $25^{th}$ percentile value. For example, weights can be 0.15 for resource, 0.2 for service type, 0.1 for event types, and 0.05 for user agent. Based on possible security risks of each service-action combo, the threshold for each category is modified.
    \item If there was an error associated with the event, then add a weight (0.1). If the error is severe like a 403 or 401 type exception (refer Section \ref{sec:feature_extraction}), then add a weight (0.2).
    \item If the event is part of a burst of events for the user then add a weight (0.2). A burst of events can be defined as observing more than 50 events in the last 5 minutes for the specific account.
\end{itemize}

\subsection{Model Training}
The TGN models A and B were trained on each dataset. As described in Section \ref{sec:tgn_def}, TGN computes node embeddings using a GRU-based memory with graph attention applied to its neighbor nodes. Our implementation used a GRU cell size of 256 for the memory of each node and a two-head graph attention mechanism to compute embeddings over a neighborhood size of 16 per node. The anomaly score computation is done using a four layered MLP that converts embeddings into a value in the range (0,1) for each edge. In this research, the edges were converted into messages using a simple concatenation of inputs. When multiple edges exist between the same pair of nodes in a training batch, a mean aggregation function is applied on the messages. To generate negative samples for self-supervised learning described in Section \ref{sec:mod_self_super}, we used a value of one for K resulting in two negative samples generated for each existing edge. As described in Table \ref{tab:dataset_stats}, training for smaller datasets was run for six epochs, while for the larger ones it was run for three epochs.

\subsection{Evaluation Protocol}
To evaluate our approach, each result detected by both our new TGN-based method and the prior domain-expert rule-based system is reviewed in detail by a cloud security expert. Each log was classified as either 'Info', 'low', 'medium', 'high', or 'critical' based on increasing levels of risk from a cybersecurity point of view. Events categorized as 'Info' are those that the expert found to be of possibly no cybersecurity risk, though they could be unusual or statistically anomalous. Events classified as 'low' and more were found to be possibly malicious activity, with 'critical' events being highly suspicious and needing immediate investigation and remediation if necessary. For this study, we consider any event classified as 'Info' to be a false positive, since they do not represent a security risk.

The anomaly score thresholds of $\leq 0.5$ for model A and $\leq 0.75$ for model B were selected based on a small number of exploratory experiments on held-out data from a subset of the datasets, tuned to balance alert volume against detection of events assessed as medium severity or higher by the security expert. These thresholds are not claimed to be globally optimal and practitioners deploying this method should treat threshold selection as a calibration step.

\subsubsection{Temporal Dataset Split}
The train-test split was performed along the time axis: the first 95\% of the 60-day log window was used for training and the final 5\% (approximately the last 3 days) was reserved for evaluation. This strictly time-ordered split avoids data leakage and reflects the deployment scenario where the model is trained on historical data before being applied to future events.

This choice has important implications. Cloud access patterns are inherently temporal with usage habits, service adoption, and infrastructure changes evolving over days and weeks. A time-based split captures this drift and tests the model's ability to generalize to near-future data from the same organization, which is the operationally relevant scenario. A random split would artificially inflate performance estimates by allowing future context to inform training.

Considering the observed nature of events, it was found that in addition to the deduplication of events as described earlier, it is useful to group all events of a service within a thirty-minute window in the report generated for analyst review. Specifically, in addition to grouping all events in thirty minute buckets that have identical account, resource,  user agent, and any errors they are also grouped by event service. Note that this additional grouping by service (ignoring event name or action) is only done for reporting during inference and not for training of models or collecting statistics for the baseline. Grouping of events predicted to be possibly malicious in such a manner is found to not only greatly reduce the number of events that need to be reviewed by analysts, but also group them in a manner that provide for the investigation. For example, consider the case where an account initiates several events to copy objects from S3 buckets and then delete them. One analyst investigating these events as a group would be much more efficient than different analysts investigating parts of it. 

\subsection{Detection Results}
The results shown in Table \ref{tab:evaluation_results} are all based on such a grouping of events. For all datasets, the model flags far fewer logs as security risks for analyst review compared to the baseline. Even independent of the baseline, the proposed method generates roughly 1 grouped alert per hour in most cases, which would substantially reduce the workload of analysts who must investigate further. Even for the largest dataset,\textbf{ it identifies just about 11 logs per hour, which is a massive reduction from the $\approx$12 thousand events per hour} in the deduplicated logs in the test data. This operational alert reduction is the primary advantage demonstrated by this study. The TGN model maintains a 'memory' for each node based on its events and its neighbors' events. Using this capability, the model predicts a higher score for repeating events (including similar events) over time, even if they never occurred in the training data of the specific user or resource. Hence, it is by design expected to identify many fewer events as anomalies compared to a heuristic method such as the baseline. Because most events are benign, and alert-fatigue is the primary issue security professionals will complain about, this is a key design constraint and operational success. The thirty-minute event grouping strategy during the inference phase demonstrated significant operational value. This approach consolidated related events while preserving detection accuracy, enabling more efficient analysis of potential security incidents.

We emphasize an important caveat: because expert review was limited to events flagged by either the TGN model or the baseline, false negatives, events that are truly malicious but not surfaced by either method, cannot be estimated from this evaluation. The lower alert volume of the TGN model could reflect a better-calibrated anomaly score, but it is also possible that malicious events are missed. This trade-off is discussed further in the Limitations section.

\begin{table}[!h]
\centering
\caption{Evaluation Results by Dataset and Method}
\label{tab:evaluation_results}
\footnotesize
\setlength{\tabcolsep}{2pt}
\begin{tabular}{ccccccccc}
\toprule
\multirow{2}{*}{\textbf{Dataset}} & \multirow{2}{*}{\textbf{\begin{tabular}{c}Events\\Count\end{tabular}}} & \multirow{2}{*}{\textbf{Method}} & \multicolumn{6}{c}{\textbf{Evaluation}} \\
\cmidrule(lr){4-9}
& & & \textbf{Total} & \textbf{Info} & \textbf{Low} & \textbf{Med} & \textbf{High} & \textbf{Critical} \\
\midrule
\multirow{2}{*}{1} & \multirow{2}{*}{21489} & TGN & 96 & 87 & 6 & 3 & 0 & 0 \\
& & Baseline & 149 & 143 & 6 & 0 & 0 & 0 \\
\addlinespace
\multirow{2}{*}{2} & \multirow{2}{*}{24357} & TGN & 79 & 76 & 3 & 0 & 0 & 0 \\
& & Baseline & 127 & 127 & 0 & 0 & 0 & 0 \\
\addlinespace
\multirow{2}{*}{3} & \multirow{2}{*}{12569} & TGN & 22 & 9 & 6 & 3 & 2 & 2 \\
& & Baseline & 4 & 4 & 0 & 0 & 0 & 0 \\
\addlinespace
\multirow{2}{*}{4} & \multirow{2}{*}{831756} & TGN & 807 & 714 & 93 & 0 & 0 & 0 \\
& & Baseline & 2,578 & 2,517 & 60 & 1 & 0 & 0 \\
\addlinespace
\multirow{2}{*}{5} & \multirow{2}{*}{290730} & TGN & 62 & 56 & 6 & 0 & 0 & 0 \\
& & Baseline & 500 & 491 & 9 & 0 & 0 & 0 \\
\bottomrule
\end{tabular}
\end{table}

\subsubsection{Investigation Results}

To further understand the nature of the TGN's model successes and failures, we performed more detailed reviews of events at multiple classification levels. We review these details in summary for each type. 

TGN found a few Critical and High events that were completely missed by the baseline detector for one dataset. From the logs, these seemed to be accounts (possibly human-based on the names) using third-party GUI tools to connect to S3 as a network drive, for Windows, with abnormal user agents and destructive API calls. Mounting S3 buckets as network drives creates persistent access paths that bypass standard AWS security controls and allow for direct data exfiltration. The detection of destructive API calls, in conjunction with these access patterns, suggests high risk data manipulation activities. This discovery represents a complex security anti-pattern that manifests across multiple areas worth introspection. This notably violated fundamental security principles of least privilege and separation of duties.

The events classified as having 'medium' security risk were mostly about accounts, again possibly linked to people rather than service accounts, modifying or deleting logging policies using rare user agents that are browser-based. TGN based method found 6 such events altogether that were not identified by the baseline detector. On the other hand, the baseline identified an 'medium' event for dataset 4 where a human account was used to delete an S3 bucket manually from a browser, which was not flagged by the TGN based method. This was caused by the presence of several other S3 events between the same account and resource prior to the one flagged by the baseline. The feature engineering for the events did not differ enough to make the TGN models consider this as a different event and hence it received a high anomaly score, as it was wrongly treated as a repeat. This can be addressed in future research by applying a lower anomaly score threshold for known-risky event types.

Examples of events classified as 'low' are accounts, mostly human rather than service accounts, directly creating KMS grants or reading configuration (for example S3 bucket ownership) or modifying a backup service configuration. These are possibly not malicious, but would be good to escalate to the analysts to let them use their best judgement. For example, an employee who has quit an organization making several of such events might be problematic. The TGN based method gave significantly better results in most cases. The only exception was dataset 5, where the baseline detector found a few additional events of 'low' severity compared to the TGN based method. The root cause for this was found to be similar to be case of the missed 'medium' event for dataset 4. Note that for dataset 3, though the TGN based method found a larger number of 'Info' events, it is likely not an issue since the FPR is still extremely low and compared to the baseline the TGN method found several more highly suspicious events.

\subsection{Deployment Retrospective}

The strategic approach to feature engineering proved instrumental in the model's success. The embedding scheme for AWS services and actions, based on security risk categories (data access provision, privilege escalation, credentials exposure, and resource exposure), created a robust foundation for risk assessment. The modification to self-supervised learning, incorporating security-specific adjustments to target labels, proved crucial for real-world applicability. This innovation enabled the model to maintain high sensitivity to security-critical events while adapting to organization-specific patterns. The approach successfully identified sophisticated attack patterns, including those involving abnormal user agents and destructive API calls, even in cases where similar patterns appeared in training data. The dual-model architecture proved particularly effective in handling complex cloud environments. Model A, focusing on account-resource relationships, and Model B, analyzing account-event patterns, worked in concert to achieve high detection accuracy while minimizing false positives. This architectural decision was validated by the results in Dataset 3, where the model identified critical security events, including unauthorized S3 access patterns, that were completely missed by the baseline detector.

From a security maturity perspective, the findings reinforce how organizational practices influence detection efficacy. Organizations at higher maturity levels, with well-established IAM practices, automated provisioning, and strict controls on human access, saw fewer high-severity alerts compared to those at lower maturity levels. This aligns with the observation that many critical and high-severity findings involved human accounts performing administrative actions through non-standard interfaces - practices that would typically be restricted in more mature environments. The research demonstrates that organizations with mature security practices - including centralized identity management, automated provisioning, and strict controls on administrative access - create an environment where anomalous activities are more clearly distinguishable. This suggests the GNN approach provides compounding benefits as organizations progress in their security maturity journey, making it particularly valuable for enterprises committed to advancing their security capabilities.

\subsection{Limitations Of Evaluation} \label{sec:limits}

This paper describes research of an industrial application conducted under real-world operational constraints. Readers should interpret the findings in light of the following limitations.

\begin{itemize}
    \item Evaluation covers only flagged events- Expert review was limited to events flagged by either the TGN model or the rule-based baseline. Events not flagged by either method were not reviewed. As a result, false negatives cannot be estimated, and it is not possible to determine whether either method misses genuine security incidents. The apparent advantage in reduced alert volume could reflect a more accurate model, a more conservative anomaly score threshold, or both. Any deployment of this approach should incorporate periodic manual auditing of a random sample of non-flagged events to assess false negative rates in the target environment.
    \item Evaluation limitations due to data retention periods and labeling cost- Domain expertise to do exhaustive labeling of a test set of events, is very expensive and time consuming. In addition, due to contractual and legal data retention requirements such datasets are not allowed to be retained for longer than a pre-defined period and all manual evaluations needs to be completed in a short time window to allow for enough time to evaluate such methods. Due to this practical limitation, this study used a single cloud security expert reviewer to assess the ground truth labels (Info/Low/Medium/High/Critical) of some of the samples. This could introduce bias in evaluation due to the subjective nature of labelling, especially for the lower risk categories of ground truth labels. 
\end{itemize}

\section{Conclusion}
This paper presented an industrial case study of a self-supervised machine learning approach that processes AWS CloudTrail logs to surface potentially anomalous events for analyst review without requiring expert-labeled training data. The GNN-based approach showed remarkable ability to distinguish between routine automated operations and potentially suspicious human activities across five organizations' environments. The experimental results demonstrate possible value for organizations where established security programs seek to enhance their detection capabilities while reducing analyst workload. The ease of maintenance of the models, due to their self-supervised nature, and an average inference time on CPU of 1 ms, make this approach operationally practical. The model's ability to update its memory during inference allows it to adapt to organizational changes without frequent retraining, which is a useful property for operational deployment. In addition, due to the simple nature of the features extracted from the events, the strategy described in this paper is expected to be applicable to other cloud provider logs.

In our evaluation, the proposed approach reduced the number of grouped alerts an analyst would need to investigate to approximately 1 per hour in most cases — a small fraction of the average event volume. Compared to a statistical rule-based baseline, the TGN-based method surfaced more events classified as medium severity or higher by the security expert in most of the evaluated datasets. Future research will include development of a protocol for estimation of false negative rates in operational deployments, extending the evaluation to other cloud providers, and conducting a comparison against established unsupervised baselines. Theoretical extensions include the investigation of alternative graph structures for security event modeling, the development of advanced temporal pattern recognition mechanisms, and the exploration of transfer learning applications in security contexts. 

\bibliographystyle{IEEEtran}
\bibliography{references}

\end{document}